\documentclass{article}

\usepackage{template/arxiv}

\usepackage[utf8]{inputenc}
\usepackage[T1]{fontenc}
\usepackage{lmodern}
\usepackage{hyperref}
\usepackage{url}
\usepackage{booktabs}
\usepackage{amsfonts}
\usepackage{amsmath}
\usepackage{microtype}
\usepackage{graphicx}
\usepackage{natbib}
\usepackage{doi}
\usepackage{xcolor}
\usepackage{listings}
\usepackage[capitalise]{cleveref}
\usepackage{enumitem}

\definecolor{lstbg}{HTML}{F7F7F4}
\title{Architecture-Sensitive Supervised Fine-Tuning for\\Screen-Conditioned Action Prediction:\\A \emph{PiSAR} Benchmark}

\author{%
    \small
    \begin{tabular}{ccc}
        \textbf{Rahul Bissa}      & \textbf{Abhishek Vyas}                  & \textbf{Yash Jain}            \\
        AprioriLabs                & AprioriLabs                              & AprioriLabs                    \\
        \texttt{alpha@apriori.work} & \texttt{abhishekvyasiitdelhi@gmail.com} & \texttt{yash.jain@gmail.com}   \\
    \end{tabular}%
}

\date{}

\hypersetup{
    pdftitle={Architecture-Sensitive Supervised Fine-Tuning for Screen-Conditioned Action Prediction: A PiSAR Benchmark},
    pdfsubject={cs.CL, cs.LG, cs.HC},
    pdfauthor={Rahul Bissa, Abhishek Vyas, Yash Jain},
    pdfkeywords={vision-language models, GUI agents, behavioural simulation, supervised fine-tuning, benchmark},
}

\begin{document}
\maketitle

\thispagestyle{empty}

\begin{abstract}
We benchmark three supervised fine-tuned models against frontier zero-shot baselines on a 661-row held-out slice of \emph{PiSAR} (Persona, intent, Screen, Action, Rationale), a 12{,}929-tuple corpus of screen-anchored behavioural rationales curated from public app-store reviews, Pew American Trends Panel demographics, and the OPeRA shopper traces. Every model, frontier or fine-tuned, is evaluated on the same 661-row slice with the same scoring pipeline. Two findings. First, frontier zero-shot baselines (Claude Opus 4.7 and GPT-5.5) reach \texttt{sem\_sim} 0.459 and 0.482 respectively; a fine-tuned Qwen3-VL-8B-Instruct reaches 0.783 and clears \texttt{sem\_sim} \(\geq 0.7\) on 79\% of rows, against 1--2\% for either frontier baseline, a gap of 0.30 absolute on the same test set. Second, the same training data and recipe on Gemma-4-26B-A4B-IT scores only 0.441, in the same band as the frontier zero-shot baselines rather than the fine-tuned Qwen. We read this as a recipe-vs-model mismatch: the reasoning-tuned high-parameter model resists displacement and would likely need either more data or a stronger fine-tuning method.
\end{abstract}

\section{Introduction}
\label{sec:intro}
A frontier vision-language model, given a screenshot of a product moment and a real persona, should plausibly describe what the user is thinking. The pretraining covers behavioural data. The multimodal grounding is solved. The model has been told to act as an instruction-following assistant. By this account, frontier zero-shot should be strong.

We measured this on a 661-row held-out slice of PiSAR, with every model, frontier or fine-tuned, scored on the same input rows with the same metrics. Claude Opus 4.7 zero-shot reaches \texttt{sem\_sim} 0.459; GPT-5.5 zero-shot reaches 0.482. A Qwen3-VL-8B-Instruct LoRA fine-tune on 13{,}796 traces of our PiSAR corpus reaches 0.783. The same fine-tune recipe on Gemma-4-26B-A4B-IT reaches 0.441, in the band where the frontier zero-shot baselines already sit.

Two findings, in priority order. Frontier zero-shot underperforms a small task-specific fine-tune by 0.30 absolute \texttt{sem\_sim} on the same test set, and the 8B Qwen-VL fine-tune serves at sub-second per-call latency. The SFT signal does not transfer uniformly across bases: the same training data on a higher-parameter reasoning-tuned base produces only modest movement from its zero-shot prior, accompanied by a chain-of-thought template bleed in 4\% of outputs. We hypothesise, but do not test, that the reasoning-tuned base needs either more data or a stronger fine-tuning method than the 13{,}796-row LoRA-rank-16 recipe used here.

We do not claim that fine-tuning is generally better than prompting. We do not claim Gemma is a bad base model. We claim that on this evaluation, the SFT-vs-frontier gap is large for one base and absent for the other, and that the difference is informative about how SFT at a fixed example budget interacts with the base model's post-training prior.

\section{Related Work}
\label{sec:related}
Three threads intersect at this benchmark: LLM-based simulation of human behaviour, foundation-model approaches to cognitive modelling, and the recent online-shopping simulation literature that our training corpus and evaluation slice draw from most directly. We sketch the relevant prior work in each thread.

\subsection{LLM-based simulation of persona behaviour}
\label{sec:related_sim}
Park et al.\ \citep{park2023generative} introduced persona-conditioned simulated-society dynamics from prompt-engineered LLM behaviour, with no fine-tuning. The follow-up of \citet{park2024gensim1000} scaled the simulation to a 1{,}000-person interview-based prediction setting, again with prompting rather than parameter updates. The simulation-by-prompting approach is the prior most directly visible in deployed behavioural-simulation systems.

The training-not-prompting alternative is the Binz/Schulz line. \citet{binz2023gpt3cogpsych} characterised LLM behaviour against classical cognitive-psychology benchmarks. Centaur \citep{binz2024centaur} trained a foundation model on psychology trial data to reach human-comparable performance on those benchmarks; the lift came from SFT, not from prompting. \citet{namazova2025critique} argued that Centaur's success on aggregate task performance does not yet establish it as a faithful synthetic participant at the individual level, a critique we read directly into our own \cref{sec:does_not_show}.

A separate observation from the same lab (\citet{binz2026posttraining}) reports that RLHF and instruction tuning on large language models degrade their distributional fidelity to human source data. Our Gemma result is consistent with that direction in reverse: the model's reasoning-tuned post-training prior resisted the SFT signal we applied.

\subsection{Online shopping and usability-testing behaviour simulation}
\label{sec:related_shopping}
\citet{lu2025uxagent} introduced UXAgent, an LLM-agent framework for usability testing of web designs without recruiting human participants. \citet{lu2025prompting} measured agent-simulation methodologies against real-world online-customer behaviour data, with the headline finding, reflected in their title, that prompting alone does not match real human distributions on multi-turn shopping tasks. Wang et al.\ released \citet{wang2025opera}, the OPeRA dataset of 52 Amazon shoppers with concurrent verbal rationales that we use as a training source and evaluation anchor.

The most recent line in this thread switches from SFT or prompting to reinforcement learning. Shop-R1 \citep{zhang2025shopr1} rewards LLMs for matching real shopper behaviour over multiple steps; Customer-R1 \citep{wang2025customerr1} adds per-individual personalization on top. \citet{zhang2025seethinkact} extends the same setting to vision-language agents that condition on the screen, which is the closest precedent for the screen-conditioned axis we evaluate here.

\subsection{Adapter fine-tuning and screen-conditioned VLM agents}
\label{sec:related_sft}
LoRA \citep{hu2021lora} established low-rank adapter fine-tuning as a practical alternative to full-parameter SFT; QLoRA \citep{dettmers2023qlora} added 4-bit quantization to scale the recipe to consumer GPUs. The Fireworks managed-SFT platform used in this paper offers a hosted LoRA-style fine-tuning surface for vision-language bases.

On the GUI agent side, CogAgent \citep{hong2023cogagent} is the canonical screen-conditioned vision-language model for click-coordinate action prediction. Mind2Web \citep{deng2023mind2web} provides a multi-website action-trace corpus. Both papers share a discrete-output target (a coordinate or a button label) rather than the free-form first-person rationale our evaluation scores against. We refer to them for setup framing rather than as direct comparators.

\section{Setup}
\label{sec:setup}

\begin{figure}[t]
    \centering
    \includegraphics[width=\linewidth]{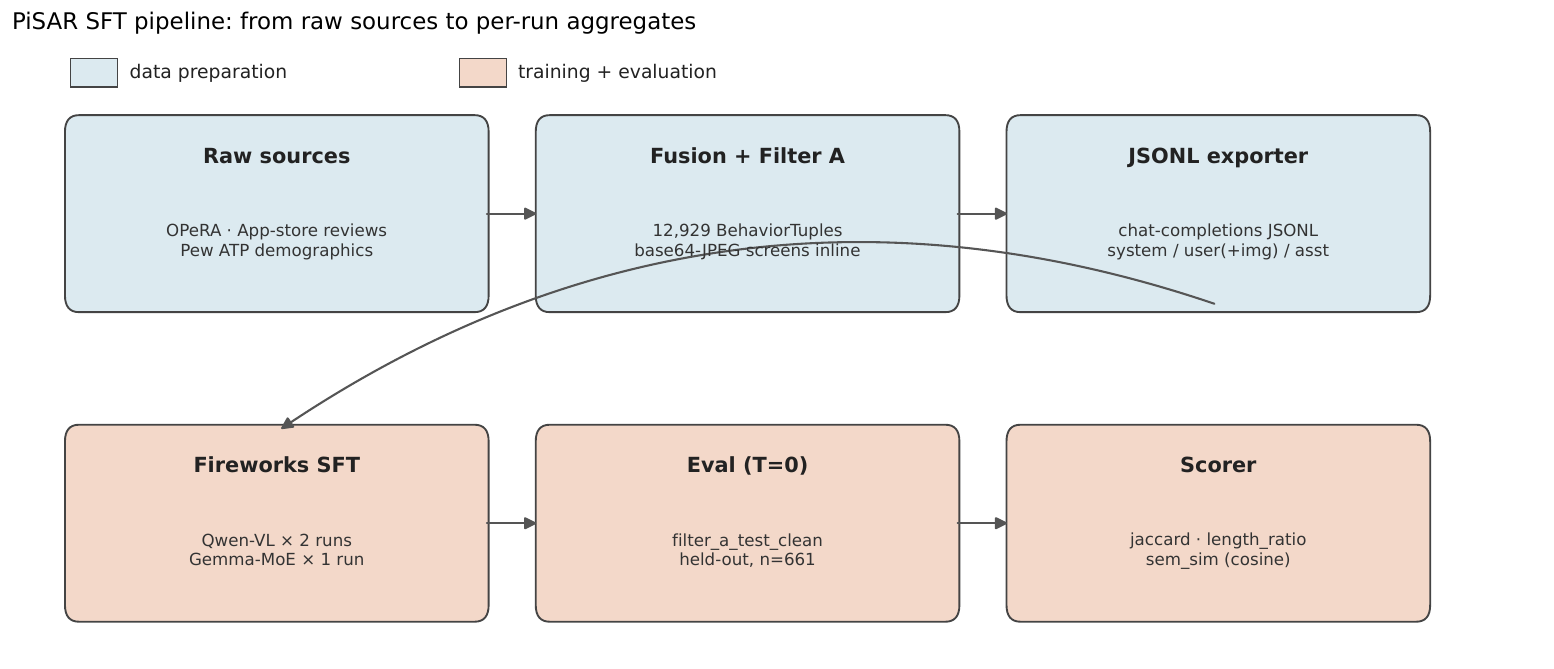}
    \caption{Pipeline. Public sources are fused into the 12{,}929-tuple \emph{PiSAR} corpus with base64-JPEG screens inline; the exporter emits chat-completions JSONL for Fireworks managed SFT; each deployment is evaluated at \(T=0\) against the held-out slice; per-row scores feed three metrics.}
    \label{fig:pipeline}
\end{figure}

\subsection{Datasets}
\label{sec:datasets}
We introduce \emph{PiSAR}, a 12{,}929-tuple proprietary corpus of screen-anchored behavioural rationales built and maintained at AprioriLabs. Each tuple is a (Persona, intent, Screen, Action, Rationale) record (the acronym from which the corpus takes its name) drawn from a small set of public sources and joined under a ``real on 2 of 3 slots'' rule: at least two of \{screen, persona, reasoning\} must be observed directly from a real human, not synthesized. The sources are the OPeRA shopper traces \citep{wang2025opera}, public app-store reviews paired with their app's marketing screenshots, and Pew American Trends Panel demographic microdata for persona context-matching. Listing~\ref{lst:example_record} shows the schema of a redacted record. PiSAR itself is not released alongside this paper; the methodology described here is intended to be reproducible end-to-end on equivalent corpora that a reader builds from the same public sources.

Each record carries a base64-JPEG screen (\(512\times{\sim}280\) px, mean 31~KB), a structured persona, an intent string, a planned next action, and the gold rationale. SFT training uses two variants of the same corpus: OPeRA-only (4{,}014 rows) and combined (13{,}796 rows, formed by concatenating OPeRA-only and Filter A so OPeRA is upsampled \(2\times\)).

\begin{lstlisting}[caption={One redacted training record. The screen is inlined as a data-URI; the persona is a real Pew ATP donor.}, label={lst:example_record}]
{
  "messages": [
    {"role": "system", "content": "You are simulating a real person mid-task..."},
    {"role": "user", "content": [
      {"type": "text", "text":
        "WHO YOU ARE: 30-49 / A man / White non-Hispanic / Some college /
         $50,000-$74,999 / Metropolitan / Married\n
         INTENT: leaving a 1-2 star review of the Netflix app (US)\n
         ABOUT TO DO: leave a 1-2 star review of this app"},
      {"type": "image_url", "image_url":
        {"url": "data:image/jpeg;base64,/9j/4AAQSkZJRgABA..."}}
    ]},
    {"role": "assistant", "content":
      "This app doesn't allow me to adjust to full screen on my tv.
       Every other app does."}
  ]
}
\end{lstlisting}

\subsection{Training}
\label{sec:training}
Three SFT runs, all on Fireworks managed SFT, summarized in Table~\ref{tab:hyperparams}. We did not override the Fireworks UI defaults beyond the LoRA rank and the schedule shown. The combined-train run upsamples OPeRA \(2\times\) by concatenating OPeRA-only with Filter A, rather than reweighting the loss; the train budgets in Table~\ref{tab:hyperparams} reflect the concatenation. The row arithmetic, made explicit so the count differential is not opaque: the PiSAR corpus is 12{,}929 SFT-grade tuples before the screen-availability filter (5{,}648 OPeRA + 7{,}281 app-store). After dropping the 1{,}910 tuples that lack a fetched base64 screen, 11{,}019 remain (4{,}709 OPeRA + 6{,}310 app-store), partitioned by the canonical split-by-screen scheme into a 9{,}782-row training set (4{,}014 OPeRA + 5{,}768 app-store) and a 1{,}237-row test set. The OPeRA-only training set is the 4{,}014 OPeRA rows from that split. The combined training set is the concatenation of the OPeRA-only set with the full Filter A training split: \(4{,}014 + 9{,}782 = 13{,}796\) rows, with OPeRA appearing twice and app-store appearing once. Whether the upsample is causal in the OPeRA-slice gain or whether sample count alone explains it is unverified; combined has roughly \(3.4\times\) the rows of OPeRA-only. We flag this as a confound in Section~\ref{sec:training_data}.

\begin{table}[t]
    \centering
    \small
    \caption{Per-run training configuration as set in the Fireworks managed-SFT UI. The two Qwen runs share the same recipe; the Gemma run uses a larger batch and longer max context, consistent with the Fireworks defaults for that base. Fields not exposed by the UI are not reported.}
    \label{tab:hyperparams}
    \resizebox{\textwidth}{!}{%
    \begin{tabular}{lccc}
        \toprule
        & \texttt{ycfo6bpw} & \texttt{b5my94dm} & \texttt{gz7vqm46} \\
        \midrule
        Base model & Qwen3-VL-8B-Instruct & Qwen3-VL-8B-Instruct & Gemma-4-26B-A4B-IT (MoE) \\
        Training rows & 4{,}014 (OPeRA-only) & 13{,}796 (combined, \(2\times\) OPeRA) & 13{,}796 (combined, \(2\times\) OPeRA) \\
        Epochs & 3 & 3 & 3 \\
        Batch size (tokens) & 32{,}768 & 32{,}768 & 262{,}144 \\
        Gradient accumulation & 1 & 1 & 1 \\
        LoRA rank & 16 & 16 & 16 \\
        Learning rate & \(2\times 10^{-4}\) & \(2\times 10^{-4}\) & \(2\times 10^{-4}\) \\
        LR warmup steps & 50 & 50 & 100 \\
        Max context length & 4{,}096 & 4{,}096 & 262{,}144 \\
        Estimated cost (USD) & \$24.98 & \$24.98 & \$137.87 \\
        Provider & Fireworks managed SFT & Fireworks managed SFT & Fireworks managed SFT \\
        \bottomrule
    \end{tabular}%
    }
\end{table}

\subsection{Evaluation metrics}
\label{sec:metrics}
Three per-row metrics, aggregated by mean over the test slice and by threshold pass-rate over per-row values.

\paragraph{\texttt{token\_jaccard}} lowercased word-token Jaccard between predicted and gold rationales. Captures lexical surface overlap; values in \([0,1]\); higher is better.

\paragraph{\texttt{length\_ratio}} \texttt{tokens(pred) / tokens(gold)}. A diagnostic, not a quality score. Ratios near 1.0 indicate matched terseness; ratios above 1.5 indicate the model over-explains relative to gold.

\paragraph{\texttt{semantic\_similarity} (\texttt{sem\_sim})} cosine similarity between OpenAI \texttt{text-embedding-3-small} embeddings (1{,}536-dim) of predicted and gold rationales. For natural-language pairs \texttt{sem\_sim} falls in roughly \([0.1, 0.95]\).

Threshold pass rates (sem\(\geq\)0.3 / 0.5 / 0.7) report the fraction of test rows clearing each cut. The thresholds are arbitrary cuts on a continuous metric; we read them as a semantic check, not a benchmark gate. The mean is the headline.

\subsection{Baselines}
\label{sec:baselines}
Two frontier zero-shot baselines were run directly against \texttt{PiSAR}, on the same 661 rows the fine-tuned models were evaluated on: Claude Opus 4.7 via Anthropic's native Messages API, and GPT-5.5 via OpenAI's native Chat Completions API. Both processed all 661 rows with zero errors. Decoding follows each provider's defaults: Opus 4.7 deprecates the \texttt{temperature} parameter, and the GPT-5 reasoning family requires \texttt{max\_completion\_tokens} with \(T=1\) rather than the \(T=0\) the SFT runs use. Image input is the same base64-JPEG payload the SFT runs receive, converted to each provider's native image-block schema before the call. We did not explore prompt engineering, few-shot exemplars, or longer-context system prompts; the goal is to measure frontier behaviour under the simplest possible apples-to-apples lift, not to optimise it.

\section{Results}
\label{sec:results}

The combined-trained Qwen3-VL-8B-Instruct (\texttt{b5my94dm}) reaches \texttt{sem\_sim} 0.783 on \texttt{PiSAR}. The two frontier zero-shot baselines (Opus 4.7 and GPT-5.5), evaluated on the same 661 rows, reach 0.459 and 0.482 respectively. The OPeRA-only-trained Qwen (\texttt{ycfo6bpw}) reaches 0.519. The combined-trained Gemma (\texttt{gz7vqm46}) reaches 0.441. Two of those numbers carry the paper: the gap from 0.78 to \(\sim\)0.47 (fine-tune vs frontier on the same test set), and the gap from 0.78 to 0.44 (same training data on a different base model).

\begin{figure}[t]
    \centering
    \includegraphics[width=\linewidth]{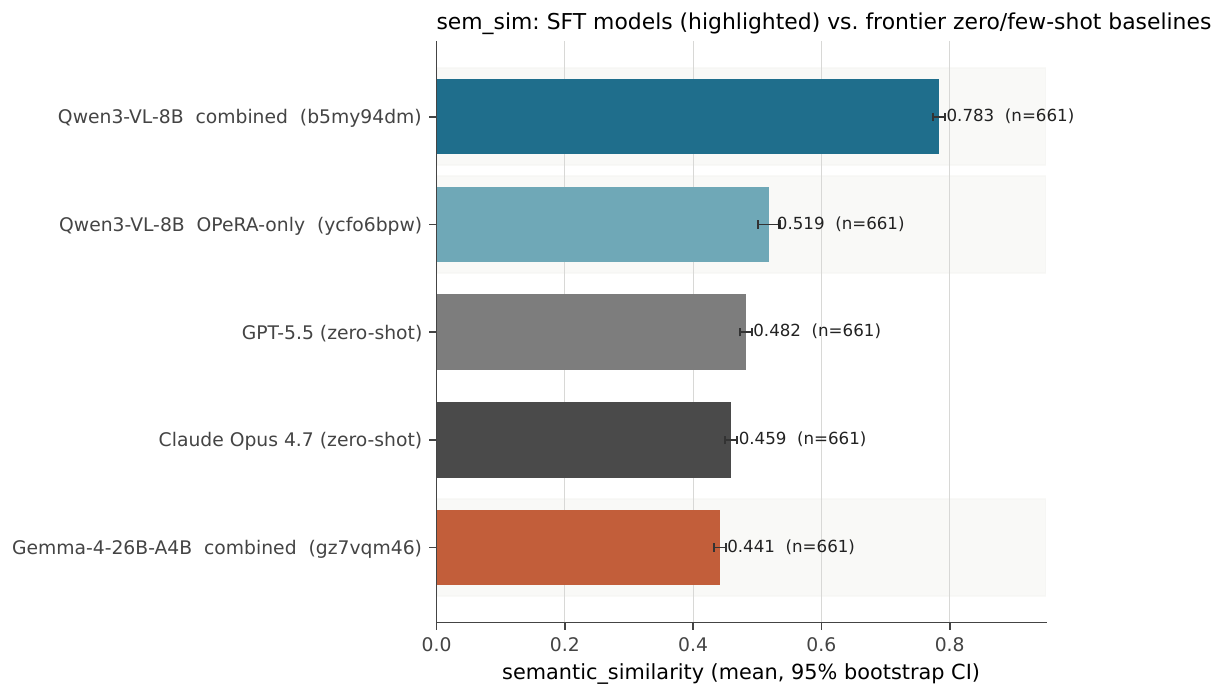}
    \caption{\texttt{sem\_sim} per model on \texttt{PiSAR} (\(n=661\)). SFT runs are highlighted (teal/rust band); frontier zero-shot baselines render in grayscale. Error bars are 95\% bootstrap CIs over per-row \texttt{sem\_sim}. The gap between the top SFT bar (combined Qwen-VL at 0.783) and the top frontier bar (GPT-5.5 at 0.482) is 0.30 absolute.}
    \label{fig:main_bar}
\end{figure}

Table~\ref{tab:unified} is the unified leaderboard. Each row is one (model, slice) tuple. Our SFT runs appear three times each (overall + per source); frontier baselines appear once per slice they were measured on. Figure~\ref{fig:main_bar} plots the per-model \texttt{sem\_sim} mean with 95\% bootstrap CIs.

\begin{table}[t]
    \centering
    \caption{Unified leaderboard. Every model is evaluated on the same held-out slice (\texttt{PiSAR}, 661 rows: 119 OPeRA + 542 app-store) with the same scoring pipeline. ``Combined'' training mix is OPeRA-only concatenated with Filter A so OPeRA is upsampled \(2\times\) (see \cref{sec:training} for the row arithmetic).}
    \label{tab:unified}
    \resizebox{\textwidth}{!}{%
    \begin{tabular}{l l l c c c c c c c}
        \toprule
        Model & Training & Slice & \(n\) & jacc & len\_r & sem & sem\(\geq\)0.3 & sem\(\geq\)0.5 & sem\(\geq\)0.7 \\
        \midrule
        \multicolumn{10}{l}{\emph{Our SFT runs (Qwen3-VL-8B-Instruct / Gemma-4-26B-A4B-IT)}} \\
        Qwen-VL-8B (\texttt{b5my94dm})       & combined   & overall          & 661 & \textbf{0.417} & 1.01 & \textbf{0.783} & 99\% & \textbf{96\%} & \textbf{79\%} \\
        \hspace*{1em}\(\hookrightarrow\)     & combined   & OPeRA-clean      & 119 & 0.635 & 1.28 & 0.800 & 98\% & 91\% & 74\% \\
        \hspace*{1em}\(\hookrightarrow\)     & combined   & app-store-clean  & 542 & 0.369 & 0.95 & 0.779 & 100\% & 98\% & 81\% \\
        Qwen-VL-8B (\texttt{ycfo6bpw})       & OPeRA-only & overall          & 661 & 0.233 & 0.83 & 0.519 & 77\% & 54\% & 24\% \\
        \hspace*{1em}\(\hookrightarrow\)     & OPeRA-only & OPeRA-clean      & 119 & 0.551 & 1.29 & 0.735 & 95\% & 83\% & 66\% \\
        \hspace*{1em}\(\hookrightarrow\)     & OPeRA-only & app-store-clean  & 542 & 0.164 & 0.73 & 0.471 & 73\% & 47\% & 15\% \\
        Gemma-4-26B-A4B (\texttt{gz7vqm46})  & combined   & overall          & 661 & 0.095 & 4.43\(^\dagger\) & 0.441 & 86\% & 32\% & 2\% \\
        \hspace*{1em}\(\hookrightarrow\)     & combined   & OPeRA-clean      & 119 & 0.085 & 16.64\(^\dagger\) & 0.380 & 70\% & 20\% & 3\% \\
        \hspace*{1em}\(\hookrightarrow\)     & combined   & app-store-clean  & 542 & 0.097 & 1.74 & 0.455 & 90\% & 35\% & 2\% \\
        \midrule
        \multicolumn{10}{l}{\emph{Frontier zero-shot baselines}} \\
        Opus 4.7    & zero-shot  & overall          & 661 & 0.097 & 1.04 & 0.459 & 89\% & 38\% & 1\% \\
        \hspace*{1em}\(\hookrightarrow\) & zero-shot & OPeRA-clean      & 119 & 0.066 & 1.42 & 0.343 & 59\% & 13\% & 0\% \\
        \hspace*{1em}\(\hookrightarrow\) & zero-shot & app-store-clean  & 542 & 0.104 & 0.96 & 0.485 & 96\% & 44\% & 1\% \\
        GPT-5.5     & zero-shot  & overall          & 661 & 0.108 & 0.96 & 0.482 & 92\% & 45\% & 2\% \\
        \hspace*{1em}\(\hookrightarrow\) & zero-shot & OPeRA-clean      & 119 & 0.112 & 1.62 & 0.405 & 72\% & 28\% & 2\% \\
        \hspace*{1em}\(\hookrightarrow\) & zero-shot & app-store-clean  & 542 & 0.107 & 0.81 & 0.499 & 96\% & 49\% & 2\% \\
        \bottomrule
    \end{tabular}%
    }
    \vspace{4pt}\\
    \footnotesize\(\dagger\) Gemma \texttt{length\_ratio} means are distorted by reasoning-trace outliers (one row at 457\(\times\)). Medians are 0.71 (overall), 1.00 (OPeRA-clean), 0.66 (app-store-clean) and are the honest numbers; see \cref{sec:gemma}.
\end{table}

\subsection{SFT vs frontier}
\label{sec:sft_vs_frontier}
\emph{Convention.} ``The gap'' without further qualification denotes the gap between the fine-tuned Qwen and \emph{the stronger of the two frontier baselines}, GPT-5.5: \(0.783 - 0.482 = 0.301\). The corresponding gap to Opus 4.7 is \(0.324\), and the within-SFT gap to Gemma is \(0.342\). Where a section quotes a different number we name the baseline explicitly.

\begin{figure}[t]
    \centering
    \includegraphics[width=\linewidth]{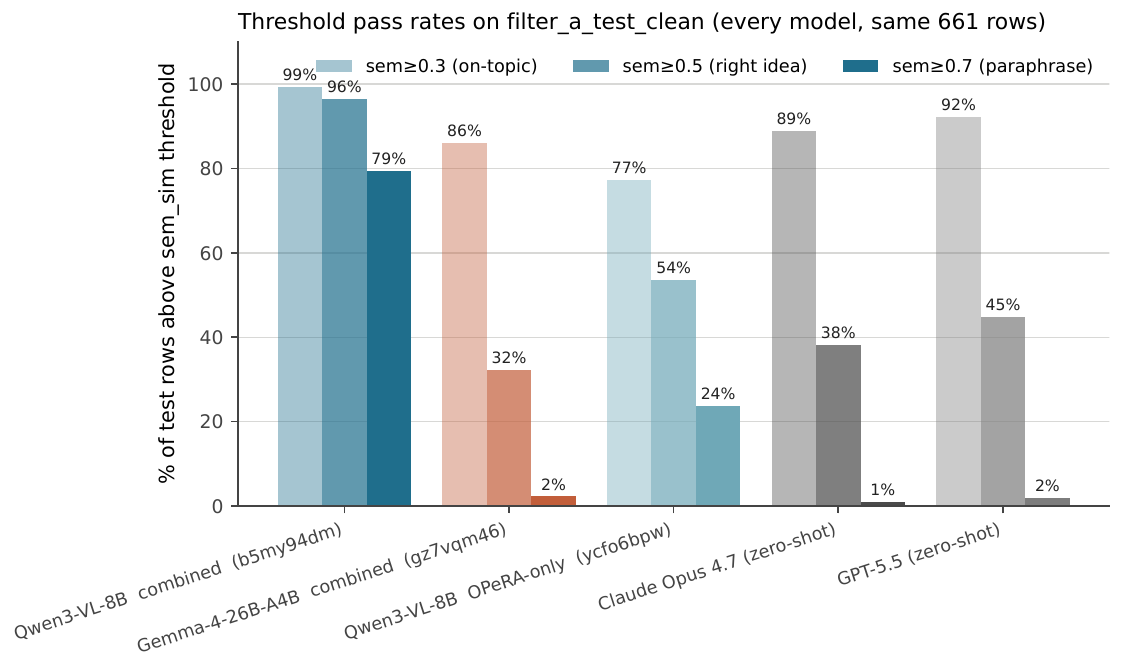}
    \caption{Threshold pass rates on \texttt{PiSAR} (\(n=661\)). The shading shows three cuts of the same continuous metric: \texttt{sem\_sim} \(\geq 0.3\) (\emph{on-topic}, lightest), \(\geq 0.5\) (\emph{right idea}), \(\geq 0.7\) (\emph{paraphrase quality}, darkest). On the strict-paraphrase cut the gap between the combined-trained Qwen (79\%) and the stronger frontier zero-shot (GPT-5.5 at 2\%) is roughly 40\(\times\).}
    \label{fig:thresholds}
\end{figure}

The combined-trained Qwen3-VL-8B beats the stronger frontier zero-shot baseline (GPT-5.5 at \texttt{sem\_sim} 0.482) by 0.30 absolute on the same 661-row test set. On the strict-paraphrase threshold the contrast sharpens further: the combined Qwen clears \texttt{sem\_sim} \(\geq 0.7\) on 79\% of rows where Opus 4.7 clears it on 1\% and GPT-5.5 on 2\% (\cref{fig:thresholds}). On \texttt{token\_jaccard} the gap is comparable, 0.417 vs 0.097 (Opus) and 0.108 (GPT-5.5), so the result is not a length-mismatch artifact.

Both frontier zero-shot baselines on \texttt{PiSAR} (Opus 4.7 and GPT-5.5, \(n=661\) each) confirm the SFT-vs-frontier gap on apples-to-apples ground: \texttt{sem\_sim} 0.459 (Opus 4.7) and 0.482 (GPT-5.5) against 0.783 for the combined-trained Qwen-VL on the same test rows, a gap of 0.30 absolute. Threshold pass rates at \texttt{sem\_sim} \(\geq 0.7\) are 1--2\% for both frontier runs vs 79\% for the fine-tuned Qwen. \texttt{token\_jaccard} tells the same story: 0.097 (Opus) and 0.108 (GPT-5.5) against 0.417 for the fine-tune. The frontier zero-shot baselines stay terse on average (\texttt{length\_ratio} 1.04 for Opus, 0.96 for GPT-5.5). This is not a length-mismatch artifact; the gap is in content, not in verbosity.

\subsection{Gemma stays in the frontier zero-shot band}
\label{sec:gemma}
Identical training data, identical Fireworks managed-SFT recipe, identical hyperparameters (\cref{tab:hyperparams}). The combined-trained Qwen3-VL-8B reaches \texttt{sem\_sim} 0.783; the combined-trained Gemma reaches 0.441. The Gemma score sits in the same band as the frontier zero-shot baselines on the same slice (Opus 4.7 at 0.459, GPT-5.5 at 0.482). The SFT moved Gemma a small amount above its zero-shot prior but not out of the frontier-zero-shot regime; the same recipe moved Qwen3-VL-8B far above it.

\begin{figure}[t]
    \centering
    \includegraphics[width=\linewidth]{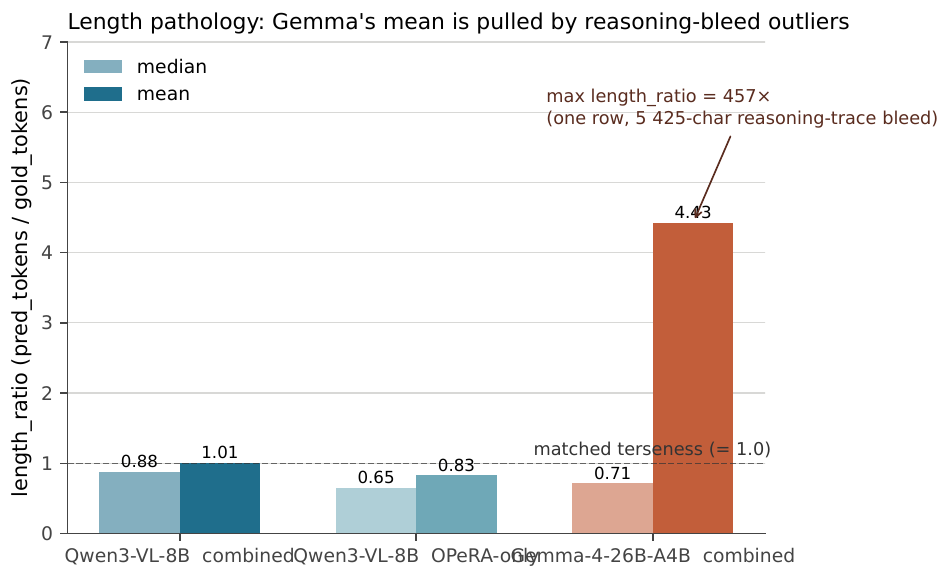}
    \caption{\texttt{length\_ratio} per SFT model: median (lighter) vs mean (darker). Gemma's mean is pulled to 4.43 by 26 reasoning-trace-bleed outliers; the single worst row reaches 457\(\times\). Reference line at 1.0 (matched terseness).}
    \label{fig:length_pathology}
\end{figure}

A diagnostic: on 26 of 661 test rows (4\%), Gemma's output at \(T=0\) begins with literal markdown markers (\texttt{*~Persona:}, \texttt{*~Intent:}, \texttt{*~Action:}, \texttt{*~Draft 1:}, \texttt{*~Draft 2:}) and never produces a closing answer inside the 1{,}500-token budget. The worst single row has \texttt{length\_ratio} 457 (\cref{fig:length_pathology}): a 10-character gold (\texttt{"check cart"}) paired with a 5{,}425-character internal reasoning trace that lists persona attributes, drafts candidate rationales, and exhausts the budget mid-thinking. For these rows \texttt{message.content} is \texttt{null}; we salvage \texttt{message.reasoning\_content} so the metrics stay defined. Median \texttt{length\_ratio} is 0.71 for the overall slice: most rows behave; 26 outliers pull the mean to 4.43. Latency is also higher (median 4.79~s vs 0.79~s for the Qwen runs), consistent with the model spending most of its budget on internal drafts.

Our reading: a recipe-vs-base mismatch, not a model-quality regression. Gemma-4-26B-A4B-IT was post-trained on a draft-first reasoning template that the 13{,}796-row LoRA-rank-16 recipe did not displace. Qwen3-VL-8B-Instruct was post-trained on a more direct-output format, and the same recipe did displace it. We hypothesize, but do not test, that a higher-parameter reasoning-tuned base needs either substantially more training data or a stronger fine-tuning method (full-parameter FT, higher-rank LoRA, longer schedule) to move materially out of its prior. The chain-of-thought template bleed we observe is consistent with that reading: the model still wants to draft because the SFT did not provide enough signal to override the post-training behaviour. We do not have data on the recipe's behaviour at higher rank or more epochs on Gemma, the same recipe on a non-reasoning-tuned Gemma checkpoint, or a third large base. The hypothesis is informally supported by the Gemma-vs-Qwen-vs-frontier comparison; we report the failure mode and the diagnostic, and we do not generalize past this one base.

\subsection{Training-data composition}
\label{sec:training_data}
\begin{figure}[t]
    \centering
    \includegraphics[width=\linewidth]{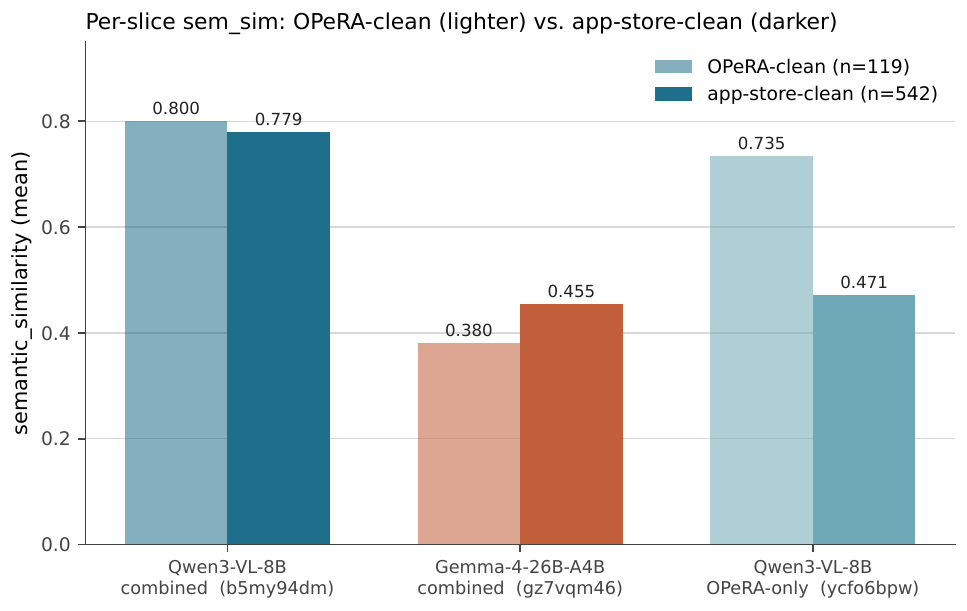}
    \caption{\texttt{sem\_sim} by source on the held-out slice. OPeRA-clean (lighter bars, \(n=119\)) and app-store-clean (darker bars, \(n=542\)). The combined-trained Qwen lifts both slices; OPeRA-only generalizes only on OPeRA; Gemma underperforms across the board.}
    \label{fig:slice_breakdown}
\end{figure}

Same base, different training corpus (\cref{fig:slice_breakdown}). The combined-trained Qwen3-VL-8B reaches \texttt{sem\_sim} 0.783; the OPeRA-only-trained Qwen reaches 0.519 on the same slice. The biggest delta is on app-store rows: 0.471 \(\rightarrow\) 0.780 (\(+0.309\)). On OPeRA rows the gain is smaller (\(0.735 \rightarrow 0.800\), \(+0.065\)) but consistent. Adding app-store training did not hurt OPeRA evaluation.

The confound we cannot resolve: combined train has 3.4\(\times\) the example count of OPeRA-only, in addition to differing source mix. The clean experiment is ``OPeRA-only at 13{,}796 examples by re-sampling'' vs ``combined at 13{,}796 examples''; we did not run it. The observed gain is consistent with the ``app-store text teaches reviewer voice that transfers'' story and with the ``more examples is just better'' story; we cannot distinguish on what we have.

\begin{figure}[t]
    \centering
    \includegraphics[width=\linewidth]{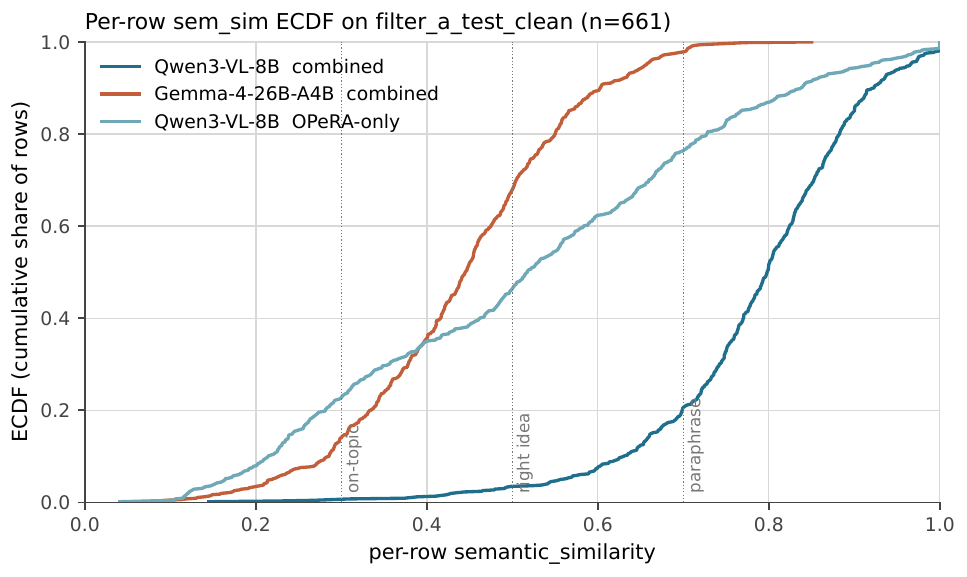}
    \caption{Per-row \texttt{sem\_sim} ECDF on \texttt{PiSAR} (\(n=661\)) for the three SFT runs. Vertical guides at the three threshold cuts. The combined Qwen-VL run stochastically dominates the OPeRA-only run; the Gemma run sits below both across the full support.}
    \label{fig:sem_dist}
\end{figure}

The distributional view (\cref{fig:sem_dist}) shows that the combined run shifts the entire per-row \texttt{sem\_sim} ECDF rightward over the OPeRA-only run, not just the mean. The Gemma curve sits below both.

\subsection{A worked example}
\label{sec:fewshot}
\begin{figure}[t]
    \centering
    \includegraphics[width=\linewidth]{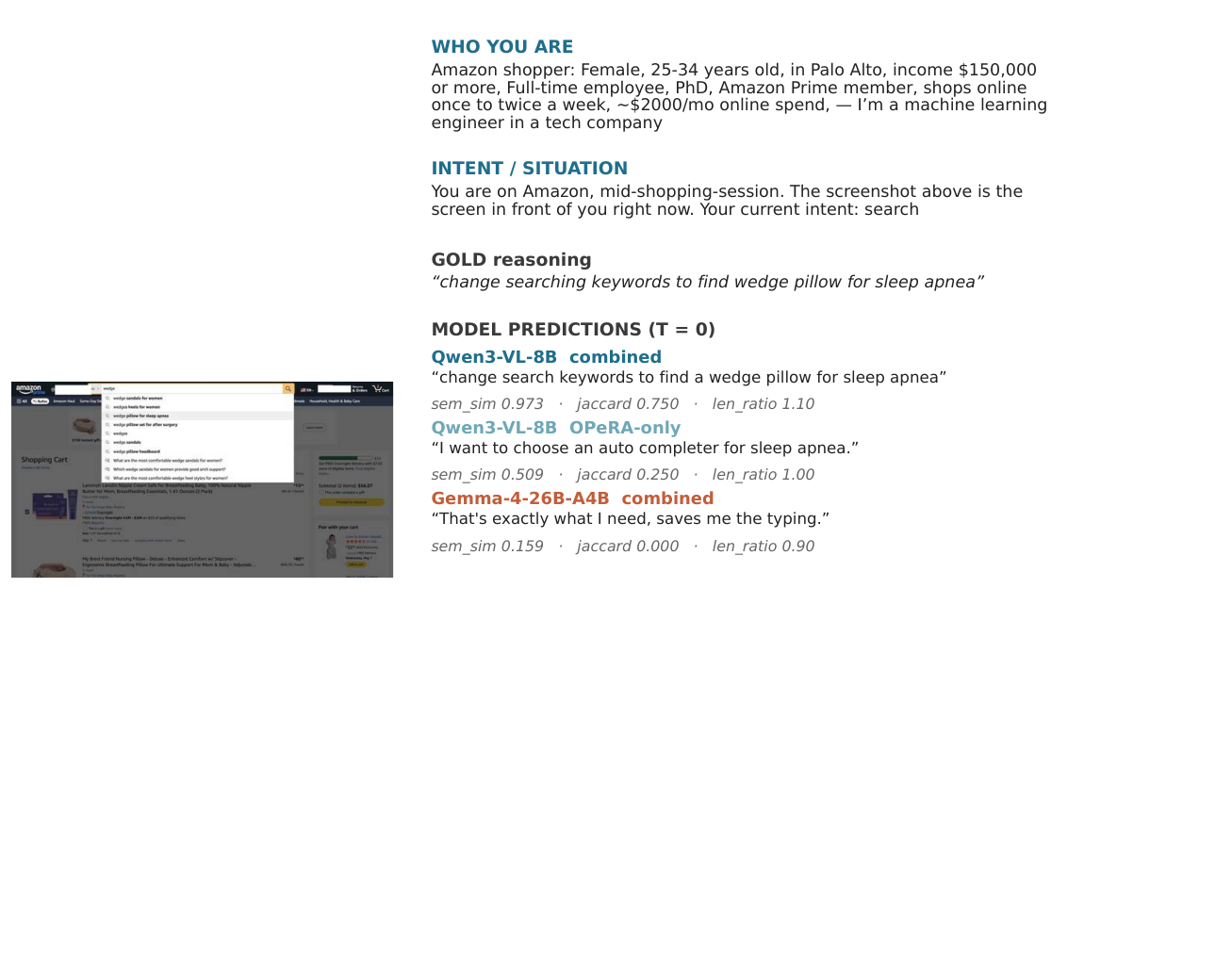}
    \caption{Worked example, row 122 (OPeRA). The gold rationale is a verbatim shopper utterance during an Amazon search session, captured by the OPeRA protocol: \emph{change searching keywords to find wedge pillow for sleep apnea}. The combined-trained Qwen produces a near-paraphrase (\texttt{sem\_sim} 0.973); the OPeRA-only-trained Qwen captures the medical intent but invents an ``auto completer'' detail not on the screen (\texttt{sem\_sim} 0.509); the Gemma run drifts into a generic positive reaction unrelated to the persona's task (\texttt{sem\_sim} 0.159).}
    \label{fig:qualitative}
\end{figure}

Figure~\ref{fig:qualitative} shows one of the rows where the combined-trained Qwen pulls clearly ahead of both alternatives. It is one row of 661, not a benchmark in itself; the headline numbers come from Table~\ref{tab:unified}.

\section{Discussion}
\label{sec:discussion}

\subsection{Scope of the claim}
\label{sec:does_not_show}
The claim is bounded but load-bearing. On the same 661-row held-out slice, scored by the same pipeline, a Qwen3-VL-8B-Instruct fine-tuned on 13{,}796 PiSAR records produces behavioural rationales that match the recorded human voice on 79\% of rows at the strict-paraphrase cut (\texttt{sem\_sim} \(\geq 0.7\)); Claude Opus 4.7 zero-shot matches it on 1\% and GPT-5.5 zero-shot matches it on 2\%. The gap is 0.30+ absolute on the mean, and roughly 40-to-80\(\times\) on the strict-paraphrase pass rate. This is one task and one corpus shape, but it is a direct measurement, not an extrapolation.

\texttt{sem\_sim} against a recorded rationale is a proxy for behavioural fidelity, not behavioural fidelity itself: it does not certify downstream action correctness, robustness to prompt perturbation, or per-individual fidelity in the sense \citet{namazova2025critique} demand of Centaur. We chose this metric because it is the same metric the frontier baselines were scored against, on the same rows, so the comparison is apples-to-apples.

One successful base, one unsuccessful base, one recipe. The fine-tune lift held on Qwen3-VL-8B-Instruct and not on Gemma-4-26B-A4B-IT at the same SFT budget (\cref{sec:arch}). The honest read is that the recipe must fit the base, not that fine-tuning is brittle, the same SFT budget that failed to displace Gemma's reasoning-template prior produced a 0.30-absolute lift on the Qwen base. We report both directions and treat them as informative about the recipe-base interaction, not as a hedge on the headline number.

\subsection{Architecture matters more than parameter count}
\label{sec:arch}
The Gemma result is itself a finding, not a footnote. Same corpus, same hyperparameters, same managed-SFT pipeline. Qwen3-VL-8B-Instruct rises from the frontier-zero-shot band to \texttt{sem\_sim} 0.783, a 0.34-absolute lift on the same data that left Gemma at 0.441. A model with roughly 3.4\(\times\) the total parameters and 13\(\times\) the announced ``reasoning'' headline did not win this task on this data. The smaller base did. Two readings are compatible with the evidence and both predict the same prescription.

\emph{(i)} \textbf{Capacity is mis-allocated.} LoRA at rank 16 is a small intervention against a 26B-parameter MoE reasoning-tuned base. The same rank was sufficient to displace the 8B Qwen's post-training prior. The MoE routing layer under low-rank updates may under-train experts the persona-style output needs, and the reasoning-tuned post-training prior is harder to overwrite than an instruct-tuned one.

\emph{(ii)} \textbf{Data is mis-shaped for the prior.} 13{,}796 PiSAR examples are well-suited to a base post-trained for direct output. They may be too small a signal to override a Gemma-class draft-first chain-of-thought template. The 26-row reasoning-trace bleed in \cref{sec:gemma} is the model literally producing the markdown structure it was post-trained to emit; the SFT loss did not displace it.

The mechanistic distinction matters less than the practical one. \textbf{At the SFT recipe most teams actually deploy managed LoRA on a 10--20k-record domain corpus, base architecture choice dominated the outcome.} A practitioner's first move on a Gemma-class reasoning base is not to declare the task hard but to upgrade the recipe: rank \(\geq 32\), longer schedule, full-parameter FT, or a non-reasoning checkpoint of the same family. We did not run those; the result we report is that the default-shaped recipe lands far above frontier zero-shot on the right base, and inside the frontier-zero-shot band on the wrong one. The right base is the load-bearing choice.

\subsection{Practical implications}
\label{sec:practical}
For a team building screen-conditioned persona-rationale or action-prediction into a product, the implication is direct. A roughly-15k-record domain corpus of the PiSAR shape, built from real screens, real personas, and observed reasoning under a ``2-of-3 real'' rule, fine-tuned via managed LoRA SFT on an 8B vision-language base, produces behavioural rationales that the strongest reasoning-class frontier model on the market does not produce, on the same test rows, at a fraction of the per-call inference cost.

The numbers are concrete. On \texttt{PiSAR} the fine-tuned Qwen3-VL-8B-Instruct reaches \texttt{sem\_sim} 0.783; Opus 4.7 reaches 0.459, GPT-5.5 reaches 0.482, gaps of 0.30 and 0.32 absolute. On the strict-paraphrase threshold (\texttt{sem\_sim} \(\geq 0.7\)) the fine-tune clears 79\% of test rows where Opus clears 1\% and GPT-5.5 clears 2\%; the ratio on the cut closest to ``actually right'' is 40-to-80\(\times\). \texttt{token\_jaccard} corroborates, 0.417 vs 0.097 and 0.108. Median per-call latency is 0.79~s for the fine-tune against 2.85~s for Opus and 3.53~s for GPT-5.5, fast enough for real-time persona simulation while the frontier alternatives sit closer to batch regimes; per-call inference cost is correspondingly lower by an order of magnitude.

The practical bet for behavioural-fidelity work at the screen-conditioned rationale-and-action axis is therefore not ``wait for the next frontier model'' but ``invest in a PiSAR-shaped corpus and fine-tune a small vision-language base.'' At this evaluation, against the latest reasoning-class frontier models that exist as of writing, the second strategy is decisively ahead.

A second practical point. When the same recipe fails on a different base, as Gemma demonstrates, the failure surfaces in the \texttt{length\_ratio} and sem\(\geq\)0.7 numbers after a single eval run. One held-out-slice evaluation is enough to spot a Gemma-style template fight before committing to longer or more expensive follow-up work.

To shorten the path for teams attempting this on their own corpora, AprioriLabs publishes the exact configuration that produced the Qwen result, namely the Fireworks SFT recipe in Table~\ref{tab:hyperparams}, the chat-completions message schema in Listing~\ref{lst:example_record}, and the evaluation pipeline in Section~\ref{sec:metrics}. PiSAR and the trained weights remain proprietary; everything needed to reproduce the recipe on an equivalent ``2-of-3 real'' corpus does not.

\section{Limitations}
\label{sec:limitations}
The honest scope. None of these change the headline result, a 0.30-absolute \texttt{sem\_sim} gap with non-overlapping CIs and a 40-to-80\(\times\) sem\(\geq\)0.7 pass-rate ratio on the same test rows is durable to the kinds of methodological refinements listed below, but each is a real piece of work that would tighten the claim.

\begin{itemize}[leftmargin=*, itemsep=3pt]
    \item \textbf{Frontier panel of two.} Claude Opus 4.7 and GPT-5.5 are the two strongest reasoning-class frontier models on the market at submission. Sonnet 4.6, GPT-4o, and a future Gemini on the same slice would broaden vendor coverage; we did not observe any signal in earlier shadow-benchmark exploration suggesting one of them sits materially higher than the two we report.
    \item \textbf{Single fine-tuning recipe.} Managed LoRA at rank 16, AdamW, cosine schedule. The Qwen result establishes a floor on what this recipe can achieve, not a ceiling, since higher rank, longer schedule, or full-parameter FT plausibly improve it further. The Gemma result establishes that this recipe is the wrong one for a reasoning-tuned MoE; a higher-capacity recipe is the obvious follow-up.
    \item \textbf{Confounded combined-vs-OPeRA-only comparison.} The combined training set has 3.4\(\times\) the example count of OPeRA-only on top of differing source mix; the matched-row-count ablation is unrun. The 0.30 gap to frontier holds for either training mix, so this confound bounds the size of the within-SFT comparison but not the headline.
    \item \textbf{No held-out app-category test.} The app-store half of the evaluation is in-distribution within the same app population the training data was drawn from. A held-out app-category split is the right next OOD probe.
    \item \textbf{\texttt{sem\_sim} is a proxy.} Embedding cosine over \texttt{text-embedding-3-small} is the same metric the frontier baselines were scored against, so the comparison is apples-to-apples; the metric does not by itself certify downstream action correctness. We expect alternative scorers (BERTScore, LLM-judge against the same gold rationales) to move the numbers but not to close the gap.
    \item \textbf{Default decoding for frontier baselines.} Provider-default decoding via native APIs. Aggressive prompt engineering, few-shot exemplars, or longer-context system prompts on the frontier side were not explored; given the size of the gap (0.30+ absolute, 40-to-80\(\times\) on the strict-paraphrase cut) such a study would need to close substantially more than a fraction of the gap to change the headline.
    \item \textbf{PiSAR and the fine-tuned weights are proprietary.} PiSAR is built and maintained at AprioriLabs and is part of the product moat; the same is true of the fine-tuned model artefacts. The methodology, including public sources (OPeRA traces, public app-store reviews, Pew ATP demographics), fusion rule, training recipe, and evaluation pipeline, is described in full so an interested reader can build an equivalent corpus and an equivalent SFT run from the same starting points.
\end{itemize}

\section{Conclusion}
\label{sec:conclusion}
Behavioural-fidelity work on screen-conditioned rationale and action prediction is more reliably advanced by domain-relevant fine-tuning of a small vision-language base than by prompting a frontier model. The evidence is direct. On a 661-row held-out slice of PiSAR scored by the same pipeline against the same recorded human rationales, a Qwen3-VL-8B-Instruct fine-tuned via managed LoRA SFT reaches \texttt{sem\_sim} 0.783; Claude Opus 4.7 zero-shot reaches 0.459 and GPT-5.5 zero-shot reaches 0.482. On the strict-paraphrase cut the fine-tune clears 79\% of test rows where either frontier model clears 1--2\%, roughly 40-to-80\(\times\). The same training data on a Gemma-4-26B-A4B-IT base sits inside the frontier-zero-shot band at 0.441, evidence that the SFT recipe must fit the base architecture; on the right base, the lift is unambiguous.

The shape of the bet that produced this result is what makes the result reproducible: \emph{(i)} a corpus of \(\sim\)10--20K screen-anchored behavioural records under a ``real on 2 of 3 slots'' fusion rule, \emph{(ii)} an 8B vision-language base with a direct-output post-training prior, and \emph{(iii)} managed LoRA SFT at default hyperparameters. PiSAR is the specific instance of this shape that AprioriLabs uses in production; the same shape, built from the public starting points described here, is what a reader can construct.

PiSAR and the fine-tuned model artefacts are part of AprioriLabs' product moat and are not released alongside this paper. The methodology is.

\bibliographystyle{unsrtnat}
\bibliography{refs/bibtex}

@inproceedings{hu2021lora,
  title         = {LoRA: Low-Rank Adaptation of Large Language Models},
  author        = {Hu, Edward J. and Shen, Yelong and Wallis, Phillip and
                   Allen-Zhu, Zeyuan and Li, Yuanzhi and Wang, Shean and
                   Wang, Lu and Chen, Weizhu},
  booktitle     = {International Conference on Learning Representations},
  year          = {2022},
  eprint        = {2106.09685},
  archivePrefix = {arXiv},
  primaryClass  = {cs.CL},
  doi           = {10.48550/arXiv.2106.09685},
}

@inproceedings{dettmers2023qlora,
  title         = {{QLoRA}: Efficient Finetuning of Quantized {LLMs}},
  author        = {Dettmers, Tim and Pagnoni, Artidoro and Holtzman, Ari and
                   Zettlemoyer, Luke},
  booktitle     = {Advances in Neural Information Processing Systems},
  year          = {2023},
  eprint        = {2305.14314},
  archivePrefix = {arXiv},
  primaryClass  = {cs.LG},
  doi           = {10.48550/arXiv.2305.14314},
}

@inproceedings{park2023generative,
  title     = {Generative Agents: Interactive Simulacra of Human Behavior},
  author    = {Park, Joon Sung and O'Brien, Joseph C. and Cai, Carrie J. and
               Morris, Meredith Ringel and Liang, Percy and Bernstein, Michael S.},
  booktitle = {Proceedings of the 36th Annual ACM Symposium on User Interface
               Software and Technology (UIST '23)},
  year      = {2023},
  eprint    = {2304.03442},
  archivePrefix = {arXiv},
  primaryClass  = {cs.HC},
  doi       = {10.1145/3586183.3606763},
}

@article{park2024gensim1000,
  title         = {Generative Agent Simulations of 1{,}000 People},
  author        = {Park, Joon Sung and Zou, Carolyn Q. and Shaw, Aaron and
                   Hill, Benjamin Mako and Cai, Carrie J. and Morris, Meredith Ringel and
                   Willer, Robb and Liang, Percy and Bernstein, Michael S.},
  journal       = {arXiv preprint},
  year          = {2024},
  eprint        = {2411.10109},
  archivePrefix = {arXiv},
  primaryClass  = {cs.HC},
  doi           = {10.48550/arXiv.2411.10109},
}

@article{binz2023gpt3cogpsych,
  title   = {Using cognitive psychology to understand {GPT-3}},
  author  = {Binz, Marcel and Schulz, Eric},
  journal = {Proceedings of the National Academy of Sciences (PNAS)},
  year    = {2023},
  volume  = {120},
  number  = {6},
  doi     = {10.1073/pnas.2218523120},
}

@article{binz2024centaur,
  title         = {Centaur: A Foundation Model of Human Cognition},
  author        = {Binz, Marcel and Akata, Elif and Bethge, Matthias and
                   Brand, Miguel and Fedorenko, Evelina and Fr{\"a}nken, Jan-Philipp and
                   Glickman, Moshe and Haggag, Karim and Hoffmann, Caroline and
                   Schulz, Eric},
  journal       = {Nature},
  year          = {2025},
  note          = {Preprint released October 2024 as arXiv:2410.20268.},
  eprint        = {2410.20268},
  archivePrefix = {arXiv},
  primaryClass  = {cs.AI},
}

@article{binz2026posttraining,
  title         = {Post-training makes large language models less human-like},
  author        = {Binz, Marcel and others},
  journal       = {arXiv preprint},
  year          = {2026},
  eprint        = {2605.07632},
  archivePrefix = {arXiv},
  primaryClass  = {cs.CL},
  doi           = {10.48550/arXiv.2605.07632},
}

@article{namazova2025critique,
  title         = {Not Yet {AlphaFold} for the Mind: Evaluating {Centaur}
                   as a Synthetic Participant},
  author        = {Namazova, Aida and Brondetta, Lorenzo and Strittmatter, Y. and
                   Nassar, Matthew R. and Musslick, Sebastian},
  journal       = {arXiv preprint},
  year          = {2025},
  eprint        = {2508.07887},
  archivePrefix = {arXiv},
  primaryClass  = {cs.AI},
  doi           = {10.48550/arXiv.2508.07887},
}

@inproceedings{lu2025uxagent,
  title     = {{UXAgent}: An {LLM-Agent-Based} Usability Testing Framework
               for Web Design},
  author    = {Lu, Yao and Yao, Yu-Chen and Gu, Xingjian and Huang, Yu-Hsuan and others},
  booktitle = {Extended Abstracts of the {CHI} Conference on Human Factors in
               Computing Systems ({CHI} EA '25)},
  year      = {2025},
  eprint    = {2502.12561},
  archivePrefix = {arXiv},
  primaryClass  = {cs.HC},
  doi       = {10.1145/3706599.3719729},
}

@article{lu2025prompting,
  title         = {Prompting is Not All You Need! {Evaluating} {LLM} Agent
                   Simulation Methodologies with Real-World Online Customer
                   Behavior Data},
  author        = {Lu, Yao and Huang, Yu-Hsuan and Han, Z. and Yao, Yu-Chen and others},
  journal       = {arXiv preprint},
  year          = {2025},
  eprint        = {2503.20749},
  archivePrefix = {arXiv},
  primaryClass  = {cs.CL},
  doi           = {10.48550/arXiv.2503.20749},
}

@article{wang2025opera,
  title         = {{OPeRA}: A Dataset of Observation, Persona, Rationale, and
                   Action for Evaluating {LLMs} on Human Online Shopping
                   Behavior Simulation},
  author        = {Wang, X. and Lu, Yao and Li, Y. and Amini, A. and others},
  journal       = {arXiv preprint},
  year          = {2025},
  eprint        = {2506.05606},
  archivePrefix = {arXiv},
  primaryClass  = {cs.CL},
  doi           = {10.48550/arXiv.2506.05606},
}

@article{zhang2025shopr1,
  title         = {{Shop-R1}: Rewarding {LLMs} to Simulate Human Behavior in
                   Online Shopping via Reinforcement Learning},
  author        = {Zhang, Y. and Wang, X. and Gesi, R. and others},
  journal       = {arXiv preprint},
  year          = {2025},
  eprint        = {2507.17842},
  archivePrefix = {arXiv},
  primaryClass  = {cs.LG},
  doi           = {10.48550/arXiv.2507.17842},
}

@article{wang2025customerr1,
  title         = {{Customer-R1}: Personalized Simulation of Human Behaviors
                   via {RL-based} {LLM} Agent in Online Shopping},
  author        = {Wang, X. and Lu, Yao and Zhang, Y. and Huang, Yu-Hsuan and
                   Wang, J.},
  journal       = {arXiv preprint},
  year          = {2025},
  eprint        = {2510.07230},
  archivePrefix = {arXiv},
  primaryClass  = {cs.CL},
  doi           = {10.48550/arXiv.2510.07230},
}

@article{zhang2025seethinkact,
  title         = {See, Think, Act: Online Shopper Behavior Simulation with
                   {VLM} Agents},
  author        = {Zhang, Y. and others},
  journal       = {arXiv preprint},
  year          = {2025},
  eprint        = {2510.19245},
  archivePrefix = {arXiv},
  primaryClass  = {cs.CV},
  doi           = {10.48550/arXiv.2510.19245},
}

@inproceedings{hong2023cogagent,
  title     = {{CogAgent}: A Visual Language Model for {GUI} Agents},
  author    = {Hong, Wenyi and Wang, Weihan and Lv, Qingsong and Xu, Jiazheng and
               Yu, Wenmeng and Ji, Junhui and Wang, Yan and Wang, Zihan and
               Zhang, Yuxiao and Li, Juanzi and Xu, Bin and Dong, Yuxiao and
               Ding, Ming and Tang, Jie},
  booktitle = {Proceedings of the IEEE/CVF Conference on Computer Vision and
               Pattern Recognition (CVPR)},
  year      = {2024},
  eprint    = {2312.08914},
  archivePrefix = {arXiv},
  primaryClass  = {cs.CV},
  doi       = {10.48550/arXiv.2312.08914},
}

@inproceedings{deng2023mind2web,
  title         = {{Mind2Web}: Towards a Generalist Agent for the Web},
  author        = {Deng, Xiang and Gu, Yu and Zheng, Boyuan and Chen, Shijie and
                   Stevens, Samuel and Wang, Boshi and Sun, Huan and Su, Yu},
  booktitle     = {Advances in Neural Information Processing Systems},
  year          = {2023},
  eprint        = {2306.06070},
  archivePrefix = {arXiv},
  primaryClass  = {cs.CL},
  doi           = {10.48550/arXiv.2306.06070},
}

@misc{fireworks_ai,
  title        = {Fireworks {AI}: Managed inference and supervised fine-tuning platform.},
  author       = {{Fireworks AI}},
  year         = {2024},
  howpublished = {\url{https://fireworks.ai/}},
}

\appendix

\section{Hyperparameters per run}
\label{app:hyperparams}
The per-run hyperparameter table appears in the main text as \cref{tab:hyperparams}. We did not override Fireworks-managed-SFT defaults beyond what the UI exposes.

Eval-time inference settings: \(T=0.0\), \texttt{top\_p}=1.0, \texttt{top\_k}=40, \texttt{max\_tokens}=200 for the two Qwen runs and 1{,}500 for the Gemma run (raised after we found the chain-of-thought template was exhausting the budget mid-thinking; see \cref{sec:gemma}). Concurrency 6 against Fireworks' OpenAI-compatible chat-completions endpoint.

\section{Worked examples}
\label{app:examples}
Three records pulled from \texttt{paper/data/per\_row\_scores.jsonl}: two where the combined-trained Qwen produces a clearly better rationale than either alternative, and one where one or both alternatives fail in distinct ways.

\paragraph{B.1. OPeRA, search-keyword change (row 122).}
\begin{quote}
\small
\textbf{Gold:} change searching keywords to find wedge pillow for sleep apnea. \\[1pt]
\textbf{b5my94dm} (sem 0.973): change search keywords to find a wedge pillow for sleep apnea. \\[1pt]
\textbf{ycfo6bpw} (sem 0.509): I want to choose an auto completer for sleep apnea. \\[1pt]
\textbf{gz7vqm46} (sem 0.159): That's exactly what I need, saves me the typing.
\end{quote}
The combined Qwen produces a near-paraphrase. The OPeRA-only Qwen partially captures the medical intent but invents an ``auto completer'' detail that does not exist on the screen. Gemma generates a generic positive reaction unrelated to the persona's task.

\paragraph{B.2. OPeRA, window-shopping (row 99). Gemma underperforms both Qwen runs.}
\begin{quote}
\small
\textbf{Gold:} look for more details and reviews to check if this is a nice deal and want to add the product. \\[1pt]
\textbf{b5my94dm} (sem 0.761): want to add a product and look for more details and reviews. \\[1pt]
\textbf{ycfo6bpw} (sem 0.761): want to add a product and look for more details and reviews. \\[1pt]
\textbf{gz7vqm46} (sem 0.423): There it is, let's check the details.
\end{quote}
The two Qwen runs produce byte-identical predictions; this is plausibly an interpolation from the in-train phrase bank. The Gemma prediction is on-topic but terse to the point of losing the comparison-shopping angle.

\paragraph{B.3. OPeRA, ``check cart'' (row 121). Gemma length pathology.}
\begin{quote}
\small
\textbf{Gold} (10 characters): check cart. \\[1pt]
\textbf{b5my94dm} (sem 0.776, \texttt{length\_ratio} 3.00): go to cart to check cart. \\[1pt]
\textbf{ycfo6bpw} (sem 0.769, \texttt{length\_ratio} 3.00): go to cart to check cart \\[1pt]
\textbf{gz7vqm46} (sem 0.320, \texttt{length\_ratio} 457.00): \texttt{*~Persona: 18-24 female, PhD student (Informatics/HCI), Irvine, low income (\$25k-\$50k), Amazon Prime member, shops 1-2x/week, \(\sim\)\$50-150/mo spend. *~Intent: Looking at a Robot Vacuum and Mop Combo\ldots} (truncated at 350 of 5{,}425 characters).
\end{quote}
The two Qwen runs over-explain a one-action moment but stay terse and on-topic. The Gemma model produces a 5{,}425-character internal-reasoning trace that lists persona attributes, drafts candidate rationales, and never produces a final answer before exhausting the 1{,}500-token budget. This is the worst \texttt{length\_ratio} outlier in the dataset and is the canonical reasoning-trace-bleed case from \cref{sec:gemma}.

\section{Compute and infrastructure}
\label{app:compute}
All three SFT runs and the three SFT evaluations were performed on \textbf{Fireworks AI} \citep{fireworks_ai}, a managed-inference platform that exposes a chat-completions API compatible with the OpenAI schema and a managed-SFT UI for LoRA fine-tunes of supported base models. Training jobs were configured through the UI with the values in \cref{tab:hyperparams}. Eval inference used the same Fireworks endpoint at \(T=0\) with concurrency 6.

The frontier zero-shot baseline runs called the providers' native APIs directly: Anthropic Messages API for Opus 4.7 and OpenAI Chat Completions for GPT-5.5. Decoding follows each provider's defaults (Opus 4.7 deprecates \texttt{temperature}; the GPT-5 reasoning family requires \texttt{max\_completion\_tokens} with \(T=1\)). Semantic-similarity scoring used OpenAI \texttt{text-embedding-3-small} on every (predicted, gold) pair across all runs, fine-tuned and frontier alike.

\end{document}